\pdfoutput=1

\documentclass[11pt]{article}

\usepackage[final]{coling}

\usepackage{times}
\usepackage{latexsym}

\usepackage[T1]{fontenc}

\usepackage[utf8]{inputenc}

\usepackage{microtype}

\usepackage{inconsolata}

\usepackage{graphicx}

\usepackage{amsmath} 
\usepackage{xcolor}
\usepackage{tabularx}
\usepackage{booktabs}
\usepackage{colortbl}
\usepackage{graphicx}
\usepackage{pifont}

\definecolor{Gray}{gray}{0.9}
\definecolor{grannysmithapple}{rgb}{0.66, 0.89, 0.63}
\definecolor{lightgreen}{HTML}{A8E4A0}

\newcommand{\newshortname}{LLaVA-1.5-TRIM}

\title{Less is More: A Simple yet Effective Token Reduction Method for Efficient Multi-modal LLMs}

\author{Dingjie Song\textsuperscript{\ding{170},\ding{171}}, Wenjun Wang\textsuperscript{\ding{170}}, Shunian Chen\textsuperscript{\ding{170}}, Xidong Wang\textsuperscript{\ding{170}}, \\ 
\textbf{Michael Guan\textsuperscript{\ding{170}}, Benyou Wang\textsuperscript{\ding{170}}\thanks{Corresponding author. Email: wangbenyou@cuhk.edu.cn}} \\
    \textsuperscript{\ding{170}}The Chinese University of Hong Kong, Shenzhen 
    \textsuperscript{\ding{171}}Lehigh University \\
    \url{https://github.com/FreedomIntelligence/TRIM/}}

\begin{document}
\maketitle

\begin{abstract}
The rapid advancement of Multimodal Large Language Models (MLLMs) has led to remarkable performances across various domains. 
However, this progress is accompanied by a substantial surge in the resource consumption of these models. 
We address this pressing issue by introducing a new approach, Token Reduction using CLIP Metric (TRIM), aimed at improving the efficiency of MLLMs without sacrificing their performance.
Inspired by human attention patterns in Visual Question Answering (VQA) tasks, TRIM presents a fresh perspective on the selection and reduction of image tokens. 
The TRIM method has been extensively tested across 12 datasets, and the results demonstrate a significant reduction in computational overhead while maintaining a consistent level of performance. 
This research marks a critical stride in efficient MLLM development, promoting greater accessibility and sustainability of high-performing models.
\end{abstract}

\section{Introduction}

The rapid development of MLLMs has demonstrated superior, and sometimes even superhuman, performance across various fields~\citep{gpt4v,geminiteam2024gemini,llava,liu2024rec,song2024milebench,ge2024mllm,song2024both}. 
However, this progress comes with a significant increase in the resources consumed by these models. 
As a result, the research community has begun to place a greater emphasis on developing efficient MLLMs~\citep{jin2024efficient,xu2024survey}. 

\begin{figure}[htbp]
    \centering
    \includegraphics[width=\linewidth]{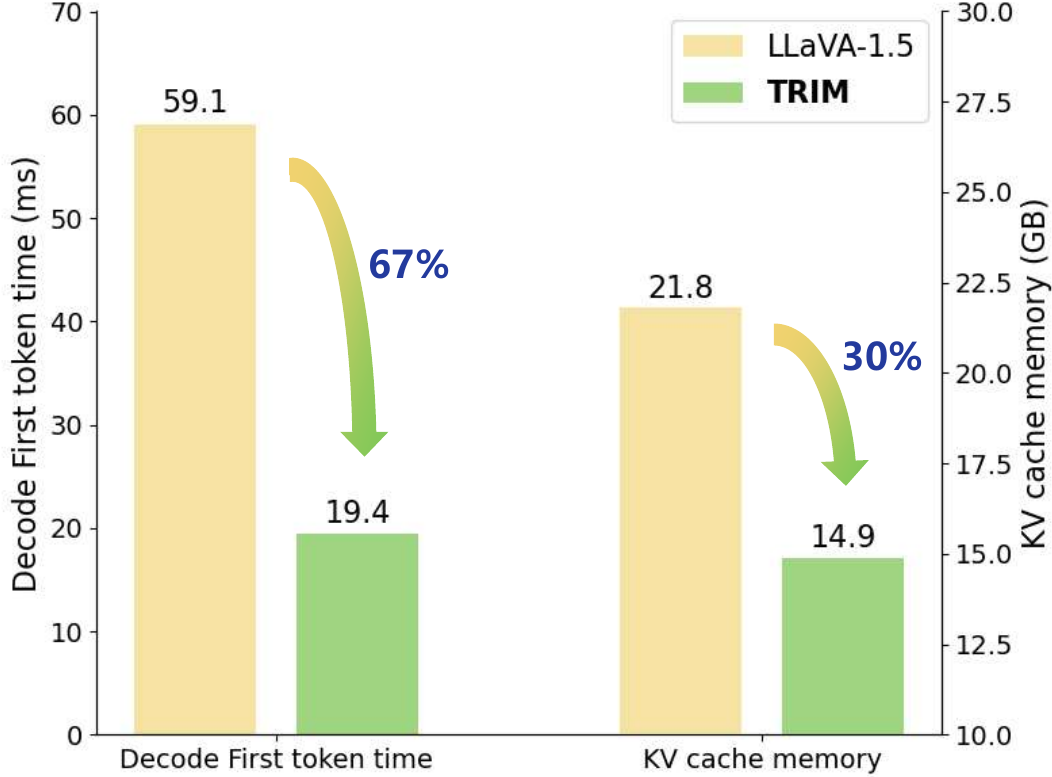}
    \caption{Comparison of First Token Decoding Time and KV Cache Memory Usage (FP16, batch size=1) for LLaVA-1.5 between the baseline and TRIM, where each image is accompanied by an average 40-token question.}
    \label{fig:fig1}
\end{figure}

Current efforts include developing lighter architectures to reduce parameters and computational complexity~\citep{lin2024moe,zhao2024cobra,chen2024allava,yuan2023tinygpt}, creating specialized components to optimize efficiency and add properties like locality~\citep{chu2024mobilevlm,cha2024honeybee}, and, notably, enhancing support for resource-intensive tasks through techniques such as visual token compression. 
Visual token compression reduces the number of tokens needed to represent visual data, thereby lowering computational and memory demands without sacrificing performance. 
This approach is particularly crucial as it enables the efficient processing of high-resolution images and videos~\citep{xu2024llava,gao2024sphinx,wang2024longllava}.

Before the MLLM era, numerous efforts aimed to reduce the number of tokens. For instance, methods like MADTP~\citep{cao2024madtp} were proposed, but they did not integrate closely enough with Large Language Models (LLMs). 
In the context of MLLMs, the only notable work is PruMerge~\citep{shang2024llavaprumerge}, which uses self-attention in vision encoder to make judgments; however, it remains a sub-optimal method for deciding which tokens to reduce.

Drawing inspiration from human attention patterns~\citep{prat2018selective} in VQA tasks, our proposed method employs the use of CLIP~\citep{radford2021learning} representations to calculate the similarity between text and image patches. Through our observations, we found that this similarity metric effectively identifies semantically relevant regions within images.

Building on this foundation, we introduce an innovative approach known as \textbf{TRIM} (\textbf{T}oken \textbf{R}eduction using CL\textbf{I}P \textbf{M}etric). 
In this method, the CLIP metric is leveraged to evaluate the significance of each image token. 
We also propose to use the Interquartile Range (IQR)~\citep{boukerche2020outlier} scoring function that adaptively selects image tokens integral to question answering.
To account for potential information loss, the selected image tokens are supplemented with an aggregated token that preserves information from the non-selected tokens. 
This methodology significantly streamlines the computational process, reducing the number of image tokens by approximately 79\%, processing time by 67\%, and memory usage by 30\% relative to the baseline, as depicted in Figure~\ref{fig:fig1}. Importantly, it achieves this efficiency while preserving performance comparable to that of the original model.

Our contributions can be summarized as follows:
\begin{itemize}
    \item We observed that the CLIP metric can effectively capture important image tokens.
    \item By leveraging the CLIP metric and the IQR scoring function, we adaptively select image tokens that are crucial for answering questions, while an aggregated token is used to retain additional image information.
    \item Extensive testing on 12 datasets demonstrates that our TRIM method significantly reduces computational overhead while maintaining consistent performance.
\end{itemize}

\section{Related Work}
Many works focus on better projecting visual information into the text embedding space.
Early work~\citep{alayrac2022flamingo} uses a perceiver resampler to integrate visual data into the language model's hidden layers. Some works~\citep{li2023blip2, zhu2023minigpt4, bai2023qwenvl, li2024finetuning,jian2024expeditedtrainingvisualconditioned} compress visual tokens to a fixed length and map them to text space using linear layers. More recent methods using the LLaVA architecture~\citep{liu2024improved, ai2024yi, wang2024cogvlm, zhu2024llavaphi, chen2024allava} simplify this by using MLP layers to map visual tokens to text space, reducing training parameters and data requirements, thus gaining popularity for their efficiency and simplicity.

However, LLaVA encounters increased computational load in multi-image scenarios due to the high number of visual tokens encoded by standard CLIP~\citep{radford2021learning, vaswani2023attention}. 
Compressing these tokens while retaining visual information is crucial. 
Though traditional CV tasks have used token merging and pruning effectively~\citep{rao2021dynamicvit,meng2021adavit,cao-etal-2023-pumer}, this approach is underexplored in MLLMs, where direct adoption of their method is more time-consuming.
LLaVA-PruMerge~\citep{shang2024llavaprumerge} recently attempted token reduction using \texttt{[CLS]} token-based similarity with sub-optimal results. In contrast, we introduce a token reduction method based on the similarity of text and visual tokens, achieving comparable performance with significantly fewer visual tokens.

\begin{figure}
    \centering
    \includegraphics[width=0.8\linewidth]{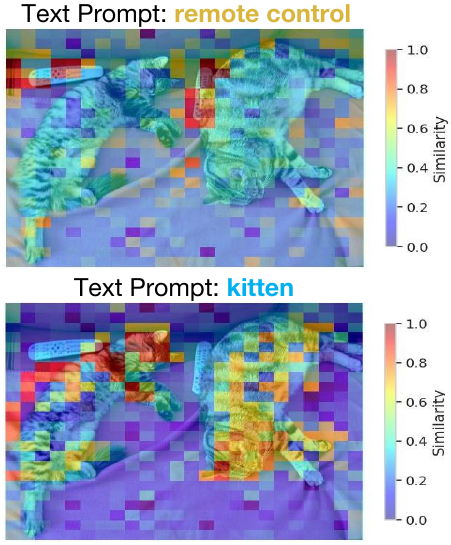}
    \caption{Similarity map between different text prompts and image patch representations within an image. Representations are extracted from the CLIP, and similarity between text and image patches is computed using dot product (CLIP metric). This demonstrates the CLIP metric's ability to capture text-image patch relationships.}
    \label{fig:obs}
\end{figure}
\section{Method}

\begin{figure*}
    \centering
    \includegraphics[width=1\linewidth]{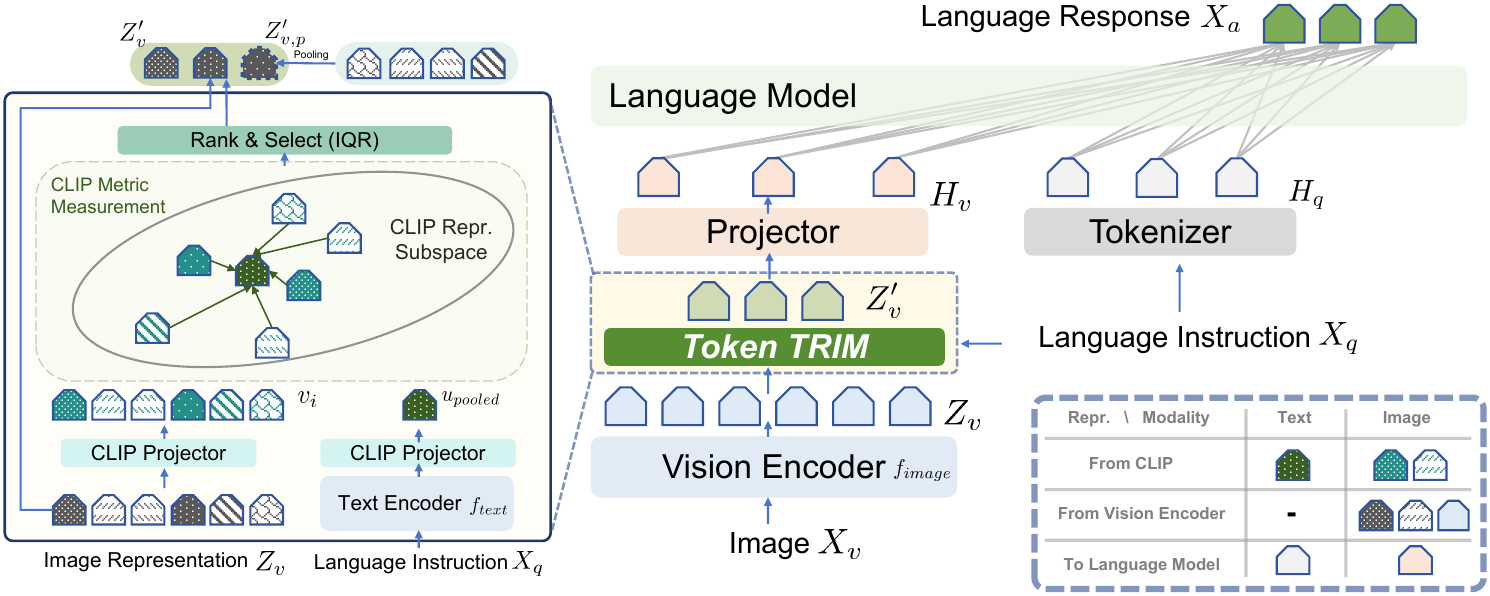}
    \caption{Overview of TRIM and LLaVA architecture. TRIM involves three steps: (1) Calculating the similarities between the text token and visual tokens; (2) Ranking and selecting the important tokens based on these similarities using an outlier detection algorithm; (3) Appending an aggregated token derived from the unselected image tokens.}
    \label{fig:main}
\end{figure*}

\subsection{Observations}

One of the challenges in token reduction is determining the importance of different tokens. 
As selective attention theory~\citep{prat2018selective} describes, selective attention in human vision prioritizes focal areas, enabling detailed processing while disregarding irrelevant information.
To simulate this attention mechanism, we need to establish a connection between the text and image patches. 
We observed that the CLIP model, during its training process, implicitly establishes such connections. 
CLIP uses contrastive learning loss to bring matching text-image pairs closer and push non-matching pairs apart. 
By leveraging these representations, we can compute and analyze the similarity between text representations and image patch representations. 
As depicted in Figure~\ref{fig:obs}, we found that by using text representations, the similarity metric effectively captured semantically relevant image patches.

\subsection{Token Reduction with TRIM}
Building upon the observations, we put forth a novel token reduction method coined as TRIM (\textbf{T}oken \textbf{R}eduction using CL\textbf{I}P \textbf{M}etric), as depicted in Figure~\ref{fig:main}, which primarily consists of three steps. 

\paragraph{Assessing Token Significance.}
First, we harness the similarity metric from CLIP to assess the significance of image tokens.
Given the feature representations extracted from the text encoder \(f_{\text{text}}\) and the image encoder \(f_{\text{image}}\), we proceed to calculate the cosine similarity between each image token \( \mathbf{v}_i \) and the pooled text representation \( \mathbf{u}_{\text{pooled}} \), derived from \texttt{[eot]} token in CLIP, as follows:

\[
S(\mathbf{v}_i, \mathbf{u}_{\text{pooled}}) = \frac{\mathbf{v}_i \cdot \mathbf{u}_{\text{pooled}}}{\|\mathbf{v}_i\| \|\mathbf{u}_{\text{pooled}}\|}
\]

Subsequently, we apply a softmax function to the calculated similarities, yielding:

\[
S_{\text{softmax}}(\mathbf{v}_i, \mathbf{u}_{\text{pooled}}) = \frac{e^{S(\mathbf{v}_i, \mathbf{u}_{\text{pooled}})}}{\sum_j e^{S(\mathbf{v}_j, \mathbf{u}_{\text{pooled}})}}
\]

This softmax score, \( S_{\text{softmax}}(\mathbf{v}_i, \mathbf{u}_{\text{pooled}}) \), effectively quantifies the significance of each image token \( \mathbf{v}_i \), thereby forming the underlying basis for the token reduction in our method.

\paragraph{Selecting Important Tokens.}
In order to determine the optimal number of image tokens to retain, we adopt the Interquartile Range (IQR) method, as suggested by \citet{shang2024llavaprumerge}. The IQR, calculated as the difference between the upper (third) \(Q3\) and lower (first) \(Q1\) quartiles of the similarity scores, is utilized as an indicator of statistical variance. We then establish a stringent similarity threshold by selecting image tokens with similarity scores that exceed the upper bound defined as \( Q3 + 1.5 \times \text{IQR} \). 
This approach ensures that only the most significant image tokens \(Z'_v\), as determined by their high similarity scores, are retained.

\paragraph{Aggregating Unselected Tokens.}
Moreover, in an effort to retain the information inherent within the unselected image tokens, we calculate an average of their representations and denote it as \(Z'_{v,p}\). This aggregated token is then appended to the selected tokens, a strategy that efficiently mitigates any potential loss of image information consequent to token reduction.
Finally, we obtain the reduced image token sequence \(Z'_{v}\).

\section{Experiment}

\begin{table*}[t!]
\centering
\scalebox{0.665}{
\begin{tabular}{l l p{5mm}>{\raggedleft\arraybackslash}p{7mm}  |p{8mm}p{8mm}p{9mm}p{7mm}p{8mm}p{8mm}p{8mm}p{8mm} p{9mm} p{8mm} p{9mm} p{13.5mm} | l}
\toprule
\rowcolor{Gray}
Method & LLM & Res. & Ratio.& VQA$^\text{v2}$ & GQA & VisWiz & SQA$^\text{I}$ & VQA$^\text{T}$ & POPE & MME & MMB & MMB$^\text{CN}$ & SEED$^\text{I}$& LLaVA$^\text{W}$ & MM-Vet  &AVG$^\dagger$ \\
\midrule
BLIP-2 & Vicuna-13B & 224 & - & 65.0 & 41.0 & 19.6 & 61.0 & 42.5 & 85.3& 1293.8 & -- & -- & 46.4 & 38.1 & 22.4  &-- \\
InstructBLIP & Vicuna-7B & 224 & - & -- & 49.2 & 34.5 & 60.5 & 50.1 & -- & -- & 36.0 & 23.7 & 53.4 & 60.9 & 26.2  &-- \\
InstructBLIP & Vicuna-13B & 224 & - & -- & 49.5 & 33.4 & 63.1 & 50.7 & 78.9 & 1212.8 & -- & -- & -- & 58.2 & 25.6  &-- \\
IDEFICS-9B & LLaMA-7B & 224 & - & 50.9 & 38.4 & 35.5 & -- & 25.9 & -- & -- & 48.2 & 25.2 & -- & -- & --  &-- \\
Qwen-VL-Chat & Qwen-7B & 448 & - & 78.2 & 57.5 & 38.9 & 68.2& 61.5& -- & 1487.5 & 60.6 & 56.7 & 58.2& -- & --  &-- \\
\midrule
\rowcolor{Gray!60}
LLaVA-1.5 & Vicuna-7B & 336 & - & 78.5& 62.0& 50.0& 66.8& 58.2& 85.9& 1510.7& 64.3& 58.3& 66.1& 65.4& 31.1 &63.5\\
 w. PruMerge& Vicuna-7B & 336 & 5.5\%& 72.0& 51.6& 43.6& 68.5& 56.0& 76.3& 1350.3& 60.9& 50.0& 50.7& 45.2&21.1&55.3\\
\rowcolor{lightgreen!20}
 w. TRIM (5\%)$^*$ & Vicuna-7B & 336 & 5\%& 71.5& 58.4& 38.4& 67.9& 49.1& 84.8& 1415.4& 63.3& 46.6& 61.8& 45.9&25.9 &\textbf{57.0}\\
  w. PruMerge+& Vicuna-7B & 336 & 25\%& 76.8& 56.4& 42.5& 68.3& 57.1& 84.0& 1462.4& 64.9& 51.8& 55.0& 51.6& 25.0&58.9\\
\rowcolor{lightgreen!60}
  w. TRIM& Vicuna-7B & 336 & 21\%& 76.4& 61.4& 48.1& 69.1& 53.7& 85.3& 1461.3& 67.4& 54.9& 65.8& 58.7& 28.0 &\textbf{61.8}\\
\midrule
\rowcolor{Gray!60}
LLaVA-1.5 & Vicuna-13B & 336 & - 
& 80.0& 63.3& 53.6& 71.6& 61.3& 85.9& 1531.3& 67.7& 63.6& 68.2& 72.5& 36.1 &66.7\\
 w. PruMerge
& Vicuna-13B & 336 & 5.5\%
& 72.8& 53.3& 48.5& 71.0& 58.4& 78.5& 1428.2& 62.3& 54.5& 54.4& 52.4&22.0&58.3\\
\rowcolor{lightgreen!20}
 w. TRIM (5\%)$^*$
& Vicuna-13B & 336 & 5\%
& 75.4& 56.0& 50.6& 70.1& 50.7& 85.2& 1337.9& 65.5& 52.4& 60.8& 45.5&24.4 &\textbf{58.6}\\
 w. PruMerge+
& Vicuna-13B & 336 & 25\%
& 77.8& 58.9& 49.7& 71.0& 58.6& 84.4& 1485.5& 65.7& 59.7& 61.4& 56.0& 28.0&62.1\\
\rowcolor{lightgreen!60}
 w. TRIM& Vicuna-13B & 336 & 21\%& 75.4& 59.0& 53.2& 72.8& 54.8& 86.3& 1438.0& 69.2& 58.3& 65.9& 57.0& 30.3 &\textbf{62.8}\\
\bottomrule
\end{tabular}
}
\caption{\textbf{Comparison with SoTA methods on 12 benchmarks.} Res, Ratio indicate input image resolution and compression ratio of image tokens, respectively. Benchmark names are abbreviated and further detailed in Appendix~\ref{sec:appendix_eval}. $^*$Select top-5\% image tokens instead of automatic selection. $^{\dagger}$Average scores across 12 datasets, with MME scaled out of 2000 points. The best performance within the same ratio range is highlighted in bold.}
\label{tab:results}
\end{table*}

\subsection{Experiment Setup}

Our experimental setup is consistent with that of LLaVA 1.5, with the key difference being that we employ our TRIM method exclusively during the instruction tuning phase. 
This approach ensures a fair comparison between our proposed method and the established baseline.
Furthermore, we perform evaluations across 12 different datasets and compare our results with those of 5 SoTA MLLMs and one related work on token reduction.
The detailed training and evaluation settings are presented in Appendix~\ref{sec:appendix_train} and Appendix~\ref{sec:appendix_eval}.

\begin{table}[t]
    \resizebox{1.0\linewidth}{!}{
    \begin{tabular}{llll}
    \toprule
    Method                 & Memory  & First Token  & Next Token  \\ 
                           & (GB)    & (ms)         & (ms)        \\ \midrule
    LLaVA-1.5 (FP16)       & 21.8    & 59.1         & 17.6        \\
    w. PruMerge       & 15.1 (69.3\%)   & 19.7 (33.3\%)      & 17.3 (98.3\%)       \\
    \rowcolor{lightgreen!60} w. TRIM                & 14.9 (68.3\%)    & 19.4 (32.8\%)         & 17.3 (98.3\%)        \\ \hline
    LLaVA-1.5 (INT8)       & 10.9   & 29.5          & 8.8         \\
    w. PruMerge       & 7.6 (69.7\%)    & 9.9 (33.6\%)         & 8.7 (98.9\%)      \\
    \rowcolor{lightgreen!60} w. TRIM                & 7.5 (68.8\%)     & 9.7 (32.9\%)          & 8.7 (98.9\%)         \\ \bottomrule
    \end{tabular}
    }
    \caption{Computation Cost Analysis. Times are measured using an NVIDIA V100 GPU, representing the hardware's theoretical peak performance (batch size=1).}
    \label{tab:eff}
\end{table}

\subsection{Main Results}
As shown in Table~\ref{tab:results}, after conducting experiments on 12 datasets, we found that despite reducing the image token count to 21\%, our method still holds a performance level comparable to LLaVA-1.5. 
Moreover, it significantly outperforms previous work such as BLIP2~\citep{li2023blip2}, InstructBLIP~\citep{dai2023instructblip}, IDEFICS-9B~\citep{idefics}, and Qwen-VL-Chat~\citep{bai2023qwenvl}. 
Our method even surpasses LLaVA-1.5 in terms of performance on the SQA$^\text{I}$ and MMB datasets. 
Compared to the previous work PruMerge, our method demonstrates superior performance across both token count (\~{}5\% and \~{}20\%) and model size (7B and 13B), despite operating with fewer image tokens. This is particularly noticeable in the GQA, POPE and MMB datasets.

\section{Analysis}

\subsection{Efficiency Analysis}

We assessed the computational efficiency using LLMViewer analysis~\cite{yuan2024llm}. In a typical scenario, a 336 $\times$ 336 pixel image processed by the CLIP model yields 576 visual tokens, alongside a 40-token text prompt. After statistical analysis, PruMerge achieves a 25\% compression rate, reducing the visual tokens to 144. 
In comparison, our method achieves a 21\% compression rate, reducing the tokens to 123. Our approach significantly accelerates model inference speed and reduces memory usage, as detailed in Table \ref{tab:eff}. Notably, the time required to generate the first token is curtailed to 32.9\% of the original, resulting in a significant acceleration in the inference process.

\subsection{Ablation Study}

\begin{table}[t]
\centering
\resizebox{0.65\linewidth}{!}{
\begin{tabular}{lcc}
\toprule
     Strategy& MMB & SEED$^\text{I}$\\
     \midrule
     LLaVA-1.5 &  64.3& 66.1\\
     \rowcolor{Gray!30}
     \textcolor{gray}{\textit{Random (21\%)}}&  \textcolor{gray}{\textit{59.3}}& \textcolor{gray}{\textit{60.2}}\\
     \rowcolor{Gray!60}
     \textcolor{gray}{\textit{Pooling (21\%)}}&  \textcolor{gray}{\textit{61.3}}& \textcolor{gray}{\textit{60.6}}\\
     \rowcolor{lightgreen!20}
     Automatic Selection &  64.1& 63.8\\
     \rowcolor{lightgreen!40}
     Aggregated Token  &  64.4& 63.8\\
     \rowcolor{lightgreen!70}
     Training& 67.4& 65.8\\
    \bottomrule
    \end{tabular}
}
\caption{Impact of TRIM strategies on performance. The first row shows baseline LLaVA-1.5 results. The second and third rows illustrate the performance at 21\% token sampling, achieved through random and linear interpolation respectively. The remaining rows exhibit incremental improvements from our strategies.}
\label{tab:abl}
\end{table}

We conducted an ablation study on the strategies proposed in our TRIM method, as shown in Table~\ref{tab:abl}. 
Initially, we analyzed the automated image token selection process based on the CLIP Metric. We compared this process to a simple linear interpolation pooling, and found that our strategy can effectively capture key information in the image, as opposed to uniform sampling (compare the second and third rows). 
The usage of an additional Aggregated token to preserve sufficient image information also results in performance gains (compare the third and fourth rows). 
Training on the basis of the TRIM strategies can further enhance results (compare the fourth and fifth rows).

\subsection{Effectiveness of TRIM Across Different Resolutions}

We investigated the effectiveness of TRIM across models with varying resolutions. 
Following the methodology of the original LLaVA authors—who used LLaMA-2-13B-Chat~\citep{touvron2023llama} and {openai/clip-vit-large-patch14}\footnote{\url{https://huggingface.co/openai/clip-vit-large-patch14}} to evaluate on LLaVA-Bench—we applied TRIM under the same conditions with a resolution of 224~\(\times\)~224 pixels. 

\begin{figure}[h]
    \centering
    \includegraphics[width=1\linewidth]{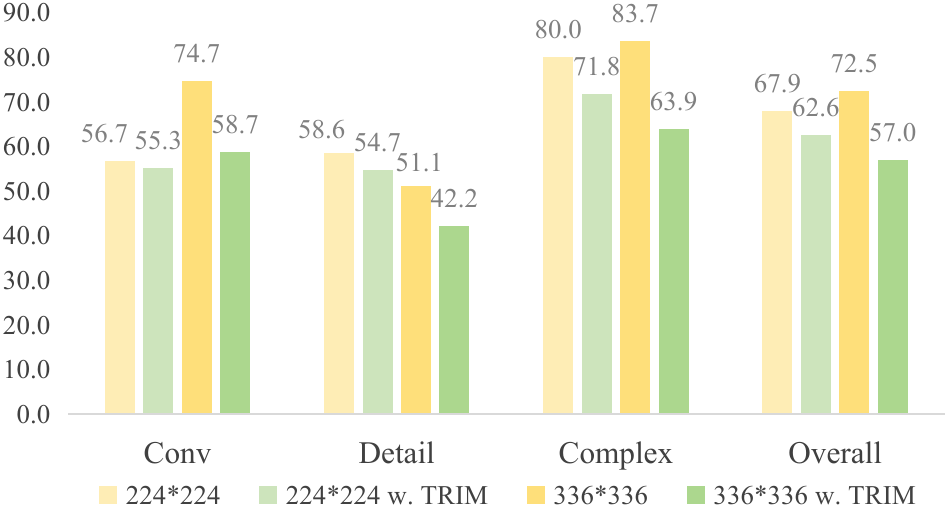}
    \caption{Comparative performance of LLaVA with and without TRIM on subsets of LLaVA-Bench across varied resolutions.}
    \label{fig:res}
\end{figure}

As shown in Figure~\ref{fig:res}, this approach yielded comparable performance, aligning with the results obtained using Vicuna-13B-v1.5\footnote{\url{https://huggingface.co/lmsys/vicuna-13b-v1.5}} and {openai/clip-vit-large-patch14-336}\footnote{\url{https://huggingface.co/openai/clip-vit-large-patch14-336}} at a resolution of 336~\(\times\)~336 pixels in Table~\ref{tab:results}.

\subsection{Effectiveness of TRIM on Video Benchmarks}

We conducted experiments on two video question-answering benchmarks: MSVD-QA~\citep{chen2011collecting} and MSRVTT-QA~\citep{xu2016msr}. For each video, we extracted frames equal to the video's duration in seconds, with a maximum of seven frames due to context length limitations. We report the accuracy and average score, with assessments performed using GPT-3.5-turbo\footnote{\url{https://platform.openai.com/docs/models\#gpt-3-5-turbo}}, as well as the first token generation time based on the average prompt length in the two datasets.

\begin{table}[h]
    \centering
    \resizebox{1.0\linewidth}{!}{
    \begin{tabular}{lccrccr}
        \toprule
        \textbf{Model} & \multicolumn{3}{c}{\textbf{MSVD-QA}} & \multicolumn{3}{c}{\textbf{MSRVTT-QA}} \\
        \cmidrule(lr){2-4} \cmidrule(lr){5-7}
        & Acc\(_\uparrow\) & Score\(_\uparrow\) & Time\(_\downarrow\) & Acc\(_\uparrow\) & Score\(_\uparrow\) & Time\(_\downarrow\) \\
        \midrule
        LLaVA-1.5-7B & 71.3 & 3.9 & 475.7 & 51.3 & 3.3 & 533.1 \\
        \rowcolor{lightgreen!30}
        ~~ w. TRIM & 67.9 & 3.9 & 76.5 & 49.7 & 3.2 & 84.2 \\
        \bottomrule
    \end{tabular}
    }
    \caption{Performance comparison of LLaVA-1.5-7B with and without TRIM on the MSVD-QA and MSRVTT-QA benchmarks. Time indicates the first token generation time in milliseconds (ms).}
    \label{tab:video-results}
\end{table}

As shown in Table~\ref{tab:video-results}, TRIM significantly reduces the first token generation time while maintaining comparable accuracy and average scores on both MSVD-QA and MSRVTT-QA benchmarks. This demonstrates that TRIM can effectively accelerate inference in video QA tasks without substantially compromising performance.

\section{Conclusion}
Our research introduced TRIM, an innovative method for reducing the image tokens in MLLMs, while maintaining performance. 
TRIM outperformed other methods, even with fewer tokens. 
Our study marks a significant step to resource-efficient MLLMs and will extend to more diverse architectures, further enhancing efficiency in the field.

\section*{Limitations}
Currently, our work is primarily limited to the widely used LLaVA architecture. 
In the future, we aim to seamlessly integrate our method into a variety of models beyond the LLaVA architecture and into different visual encoders.

\section*{Acknowledgement}

We express our gratitude to Lichao Sun for his invaluable guidance on paper structure and polish. Our appreciation is also extended to the anonymous reviewers for their constructive feedback.
This work was supported by  the Shenzhen Science and Technology Program (JCYJ20220818103001002), Shenzhen Doctoral Startup Funding (RCBS20221008093330065), Tianyuan Fund for Mathematics of National Natural Science Foundation of China (NSFC) (12326608), Shenzhen Key Laboratory of Cross-Modal Cognitive Computing (grant number ZDSYS20230626091302006), and Shenzhen Stability Science Program 2023.

\bibliography{custom}

\appendix

\section{Training Details}
\label{sec:appendix_train}

Our training method follows the instruction fine-tuning phase of LLaVA-1.5. 
The training data used is shown in Table~\ref{tab:data_mixture}, and the training hyperparameters are listed in Table~\ref{tab:hyperparameter}.
All experiments were conducted on NVIDIA A100 GPUs, with instruction-tuning for a 7B model taking approximately 8 hours and a 13B model requiring around 20 hours.

\begin{table}[ht]
\centering
\scalebox{0.67}{
\begin{tabular}{p{40mm} p{10mm}| p{50mm}}
\toprule
Data & Size & Response formatting prompts \\
\midrule
LLaVA~\cite{llava} & 158K & -- \\
ShareGPT~\cite{sharegpt} & 40K & -- \\
\midrule
VQAv2~\cite{goyal2017vqav2} & 83K & Answer the question using a single word or phrase. \\
GQA~\cite{hudson2019gqa} & 72K & \\
OKVQA~\cite{okvqa} & 9K & \\
OCRVQA~\cite{mishra2019ocrvqa} & 80K & \\
\midrule
A-OKVQA~\cite{schwenk2022okvqa} & 66K & Answer with the option's letter from the given choices directly. \\
\midrule
TextCaps~\cite{sidorov2020textcaps} & 22K & Provide a one-sentence caption for the provided image. \\
\midrule
RefCOCO & 48K & \emph{Note: randomly choose between the two formats} \\
\cite{kazemzadeh2014referitgame,mao2016generation} & & Provide a short description for this region. \\
\cmidrule{1-2}
VG~\cite{krishna2017visual} & 86K & Provide the bounding box coordinate of the region this sentence describes. \\
\midrule
Total & 665K & \\
\bottomrule
\end{tabular}
}
\caption{
Instruction-following Data Mixture of \newshortname{}.
}
\label{tab:data_mixture}
\end{table}

\begin{table}[h!]
\centering
\scalebox{0.76}{
\begin{tabular}{l |c}
\toprule
Hyperparameter & Finetune \\
\midrule
batch size & 128 \\
lr & 2e-5 \\
lr schedule & cosine decay\\
lr warmup ratio & 0.03\\
weight decay & 0\\
epoch & 1\\
optimizer & AdamW\\
DeepSpeed stage & 3 \\
\bottomrule
\end{tabular}
}
\caption{
\textbf{Hyperparameters} of \newshortname{} are the same as the original LLaVA-1.5.
}
\label{tab:hyperparameter}
\end{table}

\section{Evaluation Details}
\label{sec:appendix_eval}

\subsection{Benchmark Details}
The benchmarks employed in this study are detailed below.

VQA-v2~\cite{goyal2017vqav2};
GQA~\citep{hudson2019gqa};
VisWiz~\cite{gurari2018vizwiz};
SQA$^\text{I}$: ScienceQA-IMG~\citep{lu2022learn}; 
VQA$^\text{T}$: TextVQA~\citep{singh2019textvqa};
POPE~\cite{li2023pope};
MME (Perception)~\citep{fu2023mme};
MMB: MMBench~\citep{liu2023mmbench};
SEED$^\text{I}$: SEED-Bench-IMG~\cite{li2023seed};
LLaVA$^\text{W}$: LLaVA-Bench (In-the-Wild)~\citep{llava}; 
MM-Vet~\citep{yu2023mmvet};

\begin{itemize}

    \item VQA-v2~\cite{goyal2017vqav2} is a dataset for VQA containing 265,016 images with at least 3 questions per image and 10 ground truth answers per question. \textit{Accuracy} is used as the metric.
    
    \item VisWiz~\cite{gurari2018vizwiz} is a VQA dataset designed for assisting blind people. Each image in the dataset is accompanied by a spoken question and 10 crowdsourced answers. The challenge of the dataset includes predicting the answer to a visual question and whether a question can be answered. \textit{Accuracy} is used as the metric.
    
    \item POPE~\cite{li2023pope} is a benchmark designed to evaluate the object hallucination issue in MLLMs. The evaluation metric is the \textit{F1} score.

    \item GQA~\citep{hudson2019gqa} consists of 12,578 questions for real-world reasoning and compositional question answering. \textit{Accuracy} is used as the metric.
    
    \item MME~\citep{fu2023mme} is a benchmark with 2,374 questions spanning 14 subtasks. \textit{Accuracy} is used as the metric.
    
    \item TextVQA~\citep{singh2019textvqa} comprises 5,000 questions and \textit{Accuracy} is used as the metric.
    
    \item MM-Vet~\citep{yu2023mmvet} comprises 218 questions, each requiring multiple capabilities to solve and provided with multiple groundtruths for a flexible match. \textit{Accuracy} is adopted as the metric.
    
    \item ScienceQA~\citep{lu2022learn} contains 4,201 questions encompassing different subjects and categories. \textit{Accuracy} is adopted as the metric.
    
    \item LLaVA-Bench (In-the-Wild)~\citep{llava} contains 60 open-ended questions and uses text-based GPT-4~\citep{openai2024gpt4} as a judge to score answers in a pairwise fashion. \textit{Score Ratio} between candidate answers and anchor answers from GPT-4 is adopted as the metric.
    
    \item MMBench~\citep{liu2023mmbench} (dev set) consists of 4,329 multiple-choice questions across 20 ability dimensions, using \textit{Accuracy} under circular evaluation as the metric.
    
    \item SEED-Bench~\citep{li2023seed} (image set) comprises 14,233 multiple-choice questions across 9 dimensions. \textit{Accuracy} is adopted as the metric.

\end{itemize}

\subsection{Evaluation Prompts}
We standardize the prompt formats used for evaluation according to LLaVA-1.5, as presented in Table~\ref{tab:data_mixture_eval}.

\begin{table}[ht]
\centering
\scalebox{0.74}{
\begin{tabular}{p{38mm}| p{65mm}}
\toprule
Data & Response formatting prompts \\
\midrule
LLaVA-Bench, MM-Vet & -- \\
\midrule
VQAv2, GQA, TextVQA, MME, POPE & Answer the question using a single word or phrase. \\
\midrule
ScienceQA, MMBench, SEED-Bench & Answer with the option's letter from the given choices directly. \\
\midrule
VizWiz & When the provided information is insufficient, respond with `Unanswerable'. Answer the question using a single word or phrase. \\
\bottomrule
\end{tabular}
}
\caption{
Response format prompt for evaluation.
}
\label{tab:data_mixture_eval}
\end{table}

\end{document}